\documentclass[conference]{IEEEtran}
\IEEEoverridecommandlockouts
\usepackage{cite}
\usepackage{amsmath,amssymb,amsfonts}
\usepackage{algorithmic}
\usepackage{algorithm}
\usepackage{graphicx}
\usepackage{textcomp}
\usepackage{xcolor}
\usepackage{float}
\usepackage{booktabs}
\usepackage{multirow}
\usepackage{url}

\def\BibTeX{{\rm B\kern-.05em{\sc i\kern-.025em b}\kern-.08em
    T\kern-.1667em\lower.7ex\hbox{E}\kern-.125emX}}
\begin{document}

\title{MirrorDistill: Illumination-Aware Latent Distillation for Efficient  Low-Light Restoration\\
}

\author{
    \IEEEauthorblockN{Farida Mohsen\IEEEauthorrefmark{1}, Tala Zaim\IEEEauthorrefmark{1}, Nurul Izni Rusli\IEEEauthorrefmark{2}, Ali Al-Zawqari\IEEEauthorrefmark{3}, Ali Safa~\IEEEauthorrefmark{1}, Samir Brahim Belhaouari~\IEEEauthorrefmark{1}}
    \IEEEauthorblockA{\IEEEauthorrefmark{1}College of Science and Engineering, Hamad Bin Khalifa University, Doha, Qatar.\\ \{fmohsen, taza89388, asafa, sbelhaouari\}@hbku.edu.qa}
    \IEEEauthorblockA{\IEEEauthorrefmark{2}Faculty of Electrical Engineering \& Technology, Universiti Malaysia Perlis, Malaysia\\ nurulizni@unimap.edu.my}
    \IEEEauthorblockA{\IEEEauthorrefmark{3}ELEC Department, Vrije Universiteit Brussel, Brussels, Belgium.\\ ali.mohammed.mohammed.al-zawqari@vub.be}
}

\maketitle
\begin{abstract}
Low-light image enhancement (LLIE) is an important component of visual sensing systems operating under degraded illumination, including nighttime surveillance, autonomous navigation, remote sensing, and inspection in poorly lit industrial environments. Most LLIE methods rely on output-level reconstruction losses that supervise only the final restored image, leaving the intermediate feature recovery process weakly constrained. This paper proposes \textit{MirrorDistill}, an illumination-aware latent distillation framework that links the low-light and clean domains through feature mirroring. During training, a shared encoder and an exponential-moving-average teacher decoder process the clean reference image to generate clean-domain latent targets. These targets supervise the low-light student at two levels: raw encoder features and standardized multi-scale decoder projections. The alignment is applied layer by layer, while a proposed illumination-aware weighting scheme gives greater emphasis to underexposed regions. The teacher and reference branches are used only during training, so inference requires only the lightweight student encoder-decoder and introduces no teacher-side computational cost. Under evaluation on the standard LOL benchmarks, \textit{MirrorDistill} outperforms the state-of-the-art methods on the real-captured LOL-v2-Real set, while having the lowest compute complexity (GMACs) and while remaining competitive on the LOL-v1 and LOL-v2-Synthetic datasets. Ablation studies further show the contributions of the encoder mirror, decoder mirror, and illumination-aware weighting. Finally, we release our code as open-source for the benefit of future research. 
\end{abstract}


\section*{Supplementary Material}
The source code will be made publicly available at:
\\
\url{https://tinyurl.com/msdujxhs}

\section{Introduction}

Images acquired under low illumination are common in visual sensing systems deployed in uncontrolled environments, including nighttime surveillance, autonomous navigation, remote sensing, and inspection in poorly lit industrial facilities~\cite{10312449,10905453,9968642,Lore2017LLNet,Li2022LLIESurvey}. In these conditions, insufficient illumination causes low contrast, color distortion, amplified sensor noise, and loss of fine detail~\cite{Liang2020DeepBilateralRetinex,Lv2021DeepRetinexBilateral}. These degradations affect human interpretation and can also reduce the reliability of downstream machine-vision tasks, such as detection, tracking, and segmentation~\cite{Li2022LLIESurvey,Zheng2022SGZ,10582024}. Low-light image enhancement (LLIE) is therefore an important preprocessing component for perception pipelines that must operate under degraded illumination.

\begin{figure}[t]
  \centering
  \includegraphics[width=\columnwidth]{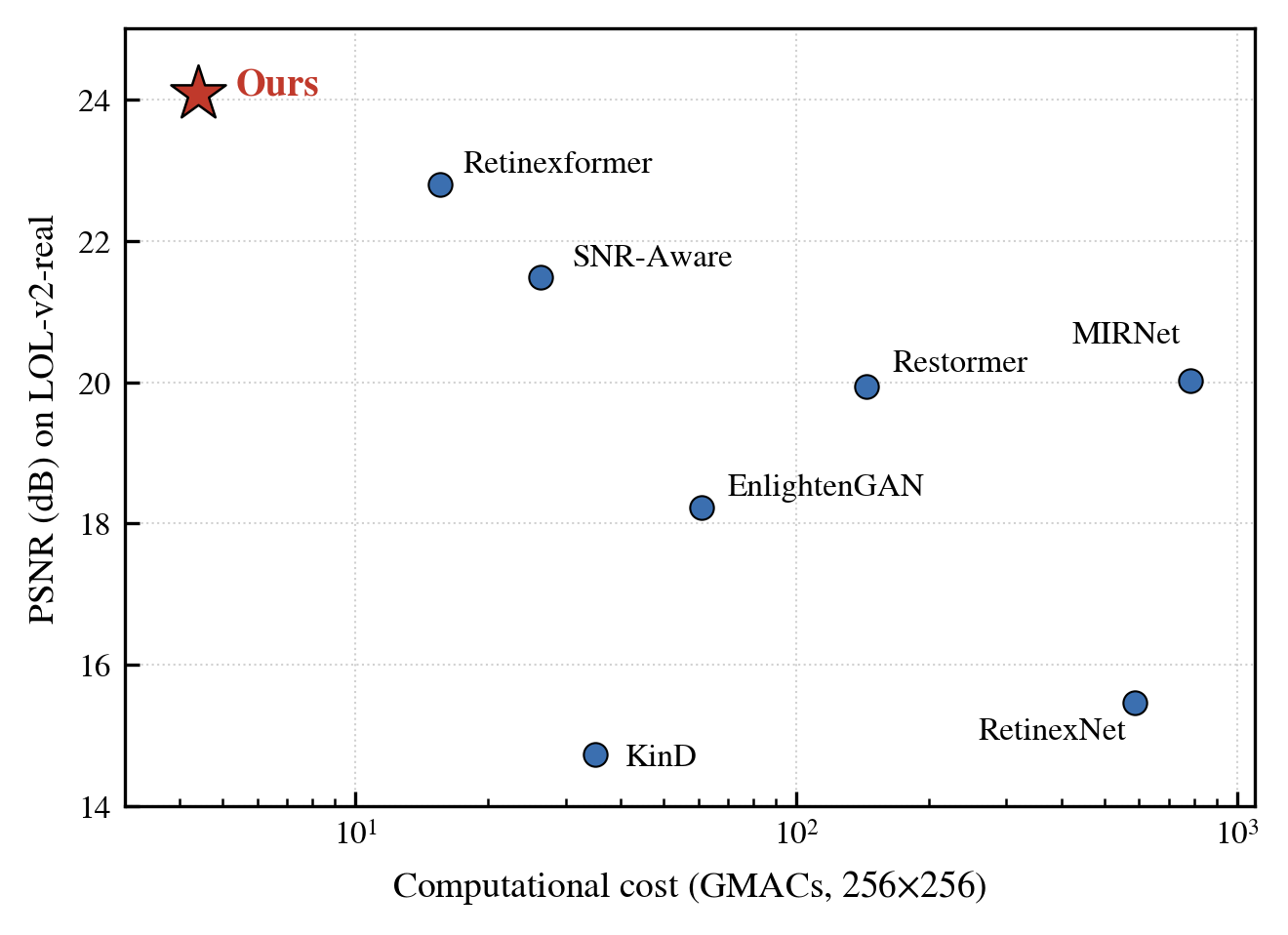}
  \caption{PSNR vs. Computational cost (GMACs, log scale) on the LOL-v2-real dataset ~\cite{yang2021sparse}. }
  \label{fig:teaser}
\end{figure}

Deep learning has substantially improved LLIE. Retinex-inspired CNN methods such as RetinexNet and KinD introduced trainable decomposition and adjustment modules for paired enhancement~\cite{Wei2018RetinexNet,Zhang2019KinD}. Other approaches reduce supervision requirements through curve estimation or unpaired adversarial learning~\cite{Li2021ZeroDCEpp,Jiang2021EnlightenGAN}. More recent restoration architectures, including transformer and attention-based models, have further improved restoration fidelity~\cite{Zamir2022Restormer,xu2022snr,Cai2023Retinexformer}. However, high restoration quality often comes with increased computational cost, including large FLOP counts, heavier memory use, and attention operations that may be difficult to deploy on resource-constrained platforms. This motivates LLIE methods that maintain competitive restoration quality while keeping inference cost predictable and low.

Beyond efficiency, paired LLIE has a supervision limitation. Most supervised LLIE methods are optimized primarily with image-level reconstruction objectives between the enhanced output and the clean reference~\cite{Li2022LLIESurvey,Wei2018RetinexNet,Zhang2019KinD}. This directly constrains the output but only indirectly guides the intermediate encoder and decoder features through which structure, illumination, and color are recovered. The problem is more pronounced in severely underexposed regions, where the input contains weak signal and the network must recover scene details from noisy and low-contrast observations.

We address these issues by proposing \textit{MirrorDistill}, an illumination-aware latent distillation framework for paired LLIE. During training, the clean reference is processed by a shared encoder and an exponential-moving-average (EMA) teacher decoder to produce clean-domain latent targets. These targets are distilled into the low-light student in a mirrored, layer-wise manner at two depths: raw encoder features and standardized multi-scale decoder projections. The alignment is weighted using an illumination map derived from the low-light input, which emphasizes underexposed regions where feature guidance is most useful. The clean branch, EMA teacher decoder, projection heads, and illumination weighting are used only during training. At inference, the deployed model is only the lightweight student encoder-decoder, with no teacher-side inference cost.

The main contributions are as follows:
\begin{enumerate}
\item We propose \textit{MirrorDistill}, an illumination-aware feature-mirroring framework that guides a low-light student using clean-domain latent targets during training.
\item We introduce a dual mirror alignment objective that supervises both raw encoder features and standardized multi-scale decoder projections, with illumination-aware weighting to focus the alignment on underexposed regions.
\item We evaluated our \textit{MirrorDistill} on three benchmark datasets namely LOL-v1, LOL-v2-Real, and LOL-v2-Synthetic. MirrorDistill achieves the best PSNR/SSIM on LOL-v2-Real among the compared methods, remains competitive on the other LOL benchmarks, and uses the lowest GMACs with the fastest measured inference time among methods with reported timing. Fig.~\ref{fig:teaser} summarizes this quality-efficiency trade-off on LOL-v2-Real.
\end{enumerate}

The remainder of this paper is organized as follows. Section~\ref{sec:related} reviews related work. Section~\ref{sec:method} presents our proposed MirrorDistill method. Section~\ref{sec:exp} reports the experimental setup, quantitative comparison, qualitative results, and ablations. Section~\ref{sec:con} concludes the paper.

\begin{figure*}[t]
  \centering
  \includegraphics[width=0.92\textwidth]{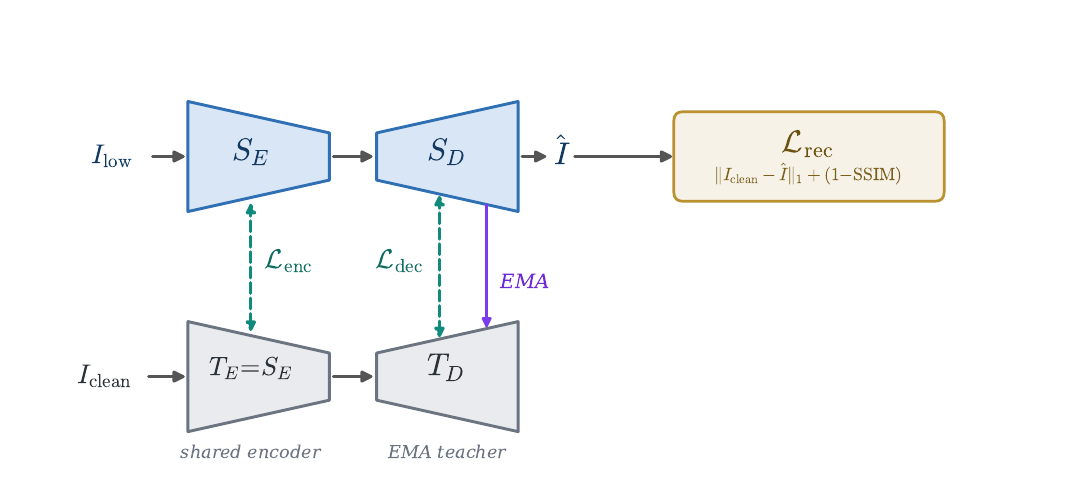}
  \caption{Overview of MirrorDistill. A shared encoder $S_E$ processes the low-light
  input (student, blue) and the clean reference (teacher, gray); the teacher decoder
  $T_D$ is an exponential moving average of the student decoder $S_D$.
  Illumination-aware mirror losses align student to teacher at the encoder
  ($\mathcal{L}_{\mathrm{enc}}$, raw features) and decoder
  ($\mathcal{L}_{\mathrm{dec}}$, per-sample standardized projections), each weighted
  by $W{=}1{+}\beta(1{-}\tilde{Y})$ ($\beta{=}0.6$) to emphasize dark regions. The
  clean branch is stop-gradient and is discarded at inference. The final deployed model during inference is
  $S_E{+}S_D$ only. 
  }
  \label{fig:arch}
\end{figure*}

\section{Related Work}
\label{sec:related}

\subsection{Low-Light Image Enhancement}
Classical LLIE methods rely on histogram equalization or on Retinex theory, which
decomposes an image into reflectance and illumination; representative model-based
approaches include LIME~\cite{Guo2016LIME} and BIMEF~\cite{Ying2017BIMEF}. Learning-
based methods now dominate. RetinexNet~\cite{Wei2018RetinexNet} introduced a deep
Retinex decomposition together with the widely used LOL dataset, and
KinD~\cite{Zhang2019KinD} improved the decomposition and adjustment stages. To reduce
the reliance on paired data, Zero-DCE++~\cite{Li2021ZeroDCEpp} reformulated
enhancement as image-specific curve estimation, and EnlightenGAN~\cite{Jiang2021EnlightenGAN}
learned from unpaired data with adversarial training. Multi-scale CNNs such as
MIRNet~\cite{Zamir2020MIRNet} 
preserve high-resolution detail while aggregating multi-scale context. More recently,
transformer-based restorers have advanced the state of the art: general models such
as Restormer~\cite{Zamir2022Restormer} and Uformer~\cite{Wang2022UFormer}, and
LLIE-specific designs such as SNR-Aware~\cite{xu2022snr}, which modulates attention by
a signal-to-noise prior, and Retinexformer~\cite{Cai2023Retinexformer}, which couples
a one-stage Retinex formulation with illumination-guided attention. Updated Retinex
models~\cite{Wu2025URetinexNetPP} and diffusion-based
enhancers~\cite{jiang2024lightendiffusion} further push fidelity. While these methods
steadily improve restoration quality, they do so with significantly increased model
and compute complexity overheads, and like most paired-based approaches, they supervise only the final
output. In contrast, our proposed \textit{MirrorDistill} targets the under-constrained latent trajectory and
keeps inference lightweight. 

\subsection{Knowledge Distillation and Feature Alignment}
Knowledge distillation  transfers knowledge from a teacher to a student~\cite{hinton2015distilling}; feature-level variants additionally align intermediate representations, as in FitNets~\cite{romero2015fitnets} and attention
transfer~\cite{zagoruyko2017attention}. The mean-teacher
framework~\cite{tarvainen2017mean} maintains the teacher as an exponential moving average of the student to provide stable targets, a strategy popular in semi-supervised
learning. Conventional KD is motivated by \emph{model compression}, where teacher and
student differ in capacity. \textit{MirrorDistill} employs distillation for a distinct purpose:
the teacher and student share the same architecture and capacity, but operate on
different input domains. The teacher processes the clean reference, while the
student processes the low-light input. This design allows the teacher to provide
clean-domain latent targets that directly guide the student's reconstruction
trajectory. Although latent mean-teacher supervision has been explored for LLIE,
as in LMT-GP~\cite{10.1007/978-3-031-73010-8_16}, our method differs in four key
aspects: it mirrors both encoder and decoder representations, aligns standardized
multi-scale decoder projections, weights the distillation signal according to
local illumination, and removes the entire distillation mechanism at inference. \emph{To sum up}, prior LLIE improves architectures and losses, while MirrorDistill adds training-time clean-domain latent guidance with no teacher-side inference cost.

\section{Proposed Method: MirrorDistill}
\label{sec:method}

\subsection{Overview and Motivation}
Most paired low-light image enhancement methods train a network by minimizing a reconstruction loss between the enhanced output and the clean reference. This supervision directly constrains the final restored image, but it provides limited guidance for the intermediate representations formed by the encoder and decoder. These representations are especially important in severely under-exposed regions, where the network must recover structure, illumination, and color from weak image evidence.

Our key observation is that the \emph{same} encoder--decoder, when applied to the
\emph{clean} reference, produces an ``ideal'' set of latent representations of the
scene. \textit{MirrorDistill} injects supervision into the latent trajectory by
distilling these clean latents into the student network that processes the
low-light input, in a mirrored, layer-by-layer fashion, at \emph{two} depths:
the raw encoder features and the standardized multi-scale decoder projections
(see Fig.~\ref{fig:arch}). Two design choices make this both effective and practical.
First, the distillation is \emph{illumination-aware}: a per-pixel weight derived
from the low-light luminance concentrates the alignment on the darkest regions,
where guidance is needed most. Second, it is \emph{training-only}: the clean pass,
the teacher branch, and the auxiliary projections are discarded at test time,
leaving a compact U-Net with \emph{no} inference-time overhead.

\subsection{Backbone and Notation}
\label{sec:backbone}
The backbone is a five-level U-Net with attention. Let $I_{\text{low}}$ and
$I_{\text{clean}}$ denote the low-light input and its clean reference. A
\emph{shared} encoder $S_E$ extracts multi-scale features from both,
\begin{equation}
\{E^i(I)\}_{i=1}^{5}=S_E(I),\qquad I\in\{I_{\text{low}},\,I_{\text{clean}}\},
\end{equation}
where each stage applies a $3{\times}3$ stride-2 convolution, batch normalization
and LeakyReLU, followed (except in the first stage) by a CBAM attention
block~\cite{woo2018cbam}, with channel widths $64,128,256,512,512$. The student
decoder $S_D$ symmetrically upsamples with transpose convolutions, BN, ReLU,
skip-concatenation and CBAM, and exposes four $1{\times}1$ projections
$\{u_j\}_{j=1}^{4}$ of its intermediate states. The enhanced image is formed as a
residual added to the input followed by a sigmoid,
\begin{equation}
\hat I=\sigma\!\big(S_D(\{E^i(I_{\text{low}})\})+I_{\text{low}}\big),
\end{equation}
which biases the network toward learning a brightening residual. Sharing $S_E$
across the two domains is what makes the encoder features directly comparable and
is the basis of the encoder mirror below.

\subsection{Clean-Teacher Latent Targets}
\label{sec:teacher}

To provide stable latent targets, we maintain a teacher decoder $T_D$ as an exponential moving average (EMA) of the student decoder $S_D$. The teacher decoder is not updated by back-propagation. Instead, its weights are updated at each training iteration as
\begin{equation}
W_{T_D} \leftarrow \mu W_{T_D} + (1-\mu) W_{S_D}, \qquad \mu = 0.999,
\label{eq:ema}
\end{equation}
where $W_{T_D}$ and $W_{S_D}$ denote the teacher and student decoder weights, respectively. The teacher decoder is initialized from $S_D$. During training, the clean reference image is passed through the shared encoder and then decoded by $T_D$ to produce clean-domain target projections ${t_j}_{j=1}^{4}$. Gradients are not propagated through the clean branch when computing the distillation losses, so these projections are treated as fixed targets for the low-light student at each iteration. Thus, the clean branch serves only as a source of latent supervision and does not receive gradient updates from the distillation objective.

\subsection{Illumination-Aware Weighting}
\label{sec:illum}
Distillation should be emphasized where the input carries the least information.
From the luminance of the low-light input,
$Y=0.299\,I_{\text{low}}^{R}+0.587\,I_{\text{low}}^{G}+0.114\,I_{\text{low}}^{B}$ (i.e., the standard luminance formula),
we form a per-image min--max normalized map $\tilde Y\in[0,1]$, resize it to a
given feature scale $s$, and define the weight
\begin{equation}
W_s=1+\beta\,(1-\tilde Y_s),\qquad \beta=0.6,
\label{eq:illum}
\end{equation}
so that the darkest pixels receive up to $1{+}\beta$ times the weight of the
brightest. This single, parameter-free prior is what makes the mirror losses
\emph{illumination-aware}.

\subsection{Illumination-Aware Mirror Losses }
\label{sec:iaml}
We distill the clean latents at two depths, in both cases weighting the per-element
discrepancy by $W_s$ from \eqref{eq:illum}.

\noindent\textbf{Encoder mirror:} The raw student encoder features
$E^i(I_{\text{low}})$ are aligned to the (stop-gradient) clean encoder features at
all five scales:
\begin{equation}
\mathcal L_{\text{enc}}=\frac{1}{5}\sum_{i=1}^{5}
\Big\langle W_i\odot\big|E^i(I_{\text{low}})-\mathrm{sg}[E^i(I_{\text{clean}})]\big|\Big\rangle,
\label{eq:enc}
\end{equation}
where $\odot$ is element-wise product and $\langle\cdot\rangle$ the mean over space
and channels.

\noindent\textbf{Decoder mirror:} The student decoder projections are aligned to the
EMA-teacher projections at all four scales. Because the student (low-light) and
teacher (clean) decoder activations differ in scale and offset, we compare them
after a per-sample standardization $\phi(z)=(z-\mu_z)/(\sigma_z+\epsilon)$ computed
over space and channels:
\begin{equation}
\mathcal L_{\text{dec}}=\frac{1}{4}\sum_{j=1}^{4}
\Big\langle W_j\odot\big|\phi(u_j)-\phi(\mathrm{sg}[t_j])\big|\Big\rangle.
\label{eq:dec}
\end{equation}
Standardization makes the decoder mirror a \emph{structural} alignment, matching activation patterns rather than absolute magnitudes, which improves the stability of cross-domain distillation.

\subsection{Training Objective}
\label{sec:obj}
The reconstruction term combines pixel and structural fidelity,
$\mathcal L_{\text{rec}}=\lVert I_{\text{clean}}-\hat I\rVert_1+\big(1-\mathrm{SSIM}(I_{\text{clean}},\hat I)\big)$,
and the full objective is
\begin{equation}
\mathcal L_{\text{total}}=\underbrace{\lVert I_{\text{clean}}-\hat I\rVert_1+\big(1-\mathrm{SSIM}\big)}_{\mathcal L_{\text{rec}}}
+\lambda_E\,\mathcal L_{\text{enc}}+\lambda_D\,\mathcal L_{\text{dec}},
\label{eq:total}
\end{equation}
with $\lambda_E=0.5$ and $\lambda_D=0.8$, empirically chosen via multiple validation trials.

\subsection{Inference}
\label{sec:opt}
At model test time, only the shared encoder $S_E$ and the student decoder $S_D$ are used:
$\hat I=\sigma(S_D(\{E^i(I_{\text{low}})\})+I_{\text{low}})$. Thus, the deployed model is the lightweight student U-Net-style encoder-decoder, with no teacher-side inference cost.

\section{Experimental Setup and Results}
\label{sec:exp}

\begin{table*}[]
\centering
\caption{\textcolor{black}{Quantitative comparison and computational complexity. PSNR/SSIM are reported on LOL-v1, LOL-v2-Real, and LOL-v2-Synthetic at full resolution evaluation. Compute complexity is reported as well (when available). Best and second-best quantitative results are shown in bold and underlined, respectively. $^{\dagger}$Runtime reported in \cite{zhou2024unveiling}.}}
\label{tab:main_native}
\begin{tabular}{l ccc cc cc cc}
\toprule
& GMACs & Params & Inference Time & \multicolumn{2}{c}{LOL-v1} & \multicolumn{2}{c}{LOL-v2-real} & \multicolumn{2}{c}{LOL-v2-syn}\\
\cmidrule(lr){5-6}\cmidrule(lr){7-8}\cmidrule(lr){9-10}
Method & (G) & (M) & (ms) & PSNR & SSIM & PSNR & SSIM & PSNR & SSIM\\
\midrule
RetinexNet~\cite{Wei2018RetinexNet}       & 587.47 & 0.84   & -- & 16.77 & 0.560 & 15.47 & 0.567 & 17.13 & 0.798\\
KinD~\cite{Zhang2019KinD}             & 34.99  & 8.02   & -- & 20.86 & 0.790 & 14.74 & 0.641 & 13.29 & 0.578\\
EnlightenGAN    & 61.01  & 114.35 & -- & 17.48 & 0.650 & 18.23 & 0.617 & 16.57 & 0.734\\
Restormer~\cite{Zamir2022Restormer}        & 144.25 & 26.13  & 104 $^{\dagger}$ & 22.43 & 0.823 & 19.94 & 0.827 & 21.41 & 0.830\\
MIRNet~\cite{Zamir2020MIRNet}           & 785.0  & 31.76  & 205 $^{\dagger}$ & 24.14 & 0.830 & 20.02 & 0.820 & 21.94 & 0.876\\
SNR-Aware~\cite{xu2022snr}         & 26.35  & 4.01   & 39 $^{\dagger}$ & \underline{24.61} & 0.842 & 21.48 & \underline{0.849} & 24.14 & 0.928\\
Retinexformer~\cite{Cai2023Retinexformer}  & 15.57 & \textbf{1.61} & 18.1 $^{\dagger}$ & \textbf{25.15} & \textbf{0.845} & \underline{22.79} & 0.840 & \textbf{25.67} & \underline{0.930}\\
\midrule
 MirrorDistill (Ours)   & \textbf{4.38} & 10.03 & \textbf{4.8} & 23.84 & \underline{0.842} & \textbf{24.08} & \textbf{0.864} & \underline{25.07} & \textbf{0.932}\\
\bottomrule
\end{tabular}

\end{table*}

\subsection{Benchmark Datasets}
We evaluate our proposed approach on three \textit{widely adopted} low-light benchmarks: \texttt{LOL-v1}, \texttt{LOL-v2-Real} and \texttt{LOL-v2-Synthetic}. First, \texttt{LOL-v1}~\cite{Wei2018RetinexNet} contains 485 training and 15 test
image pairs captured under real low-light conditions. Then,
\texttt{LOL-v2-Real}~\cite{yang2021sparse} extends the \texttt{LOL-v1} protocol with 689 training and 100 test pairs captured in real-world settings. Finally, \texttt{LOL-v2-Synthetic}~\cite{yang2021sparse} provides 900 training and 100 test pairs generated via synthetic degradation pipelines. All three datasets supply spatially aligned low-light / normal-light pairs enabling supervised training and reliable quantitative evaluation.

\begin{figure*}[h]
\centering
\includegraphics[scale=0.45]{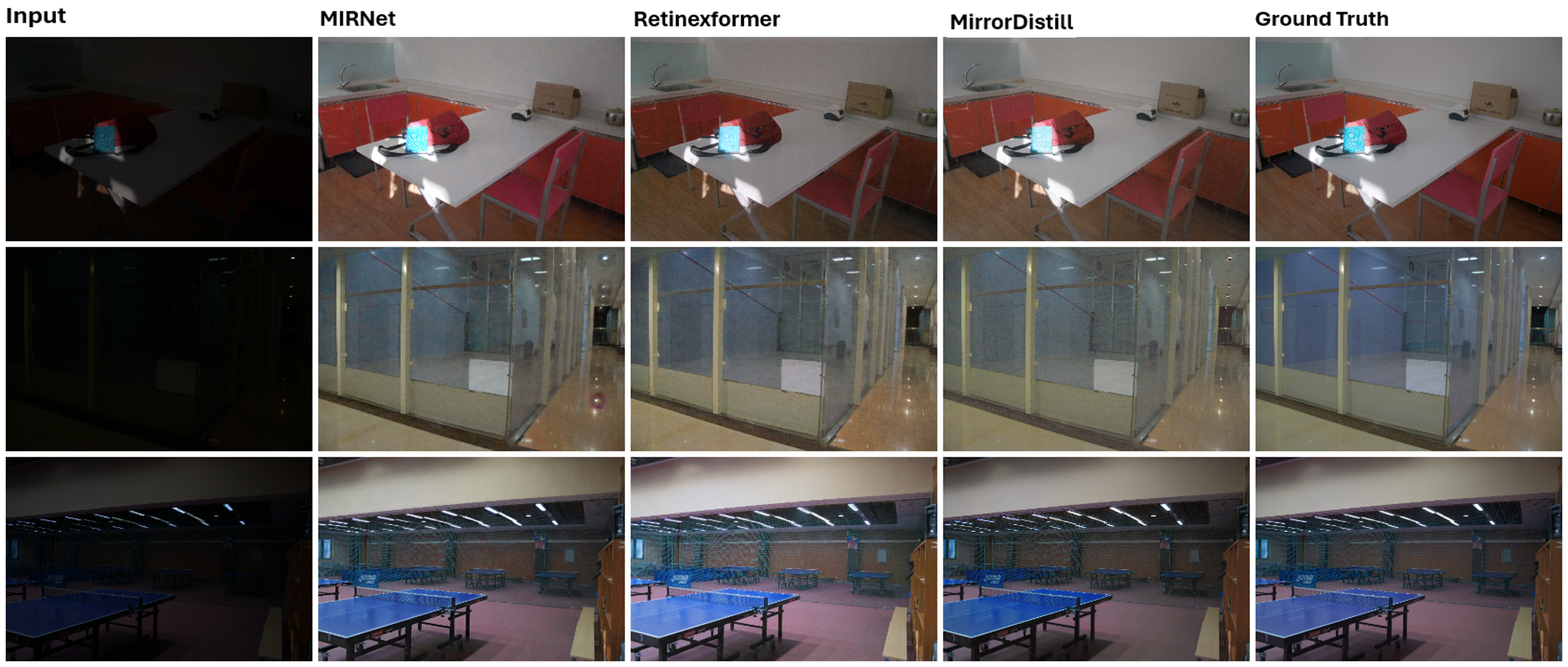}
\caption{Qualitative comparison on LOL-v2-real. }
\label{fig122}
\end{figure*}


\subsection{Implementation Details}
\label{sec:impl}
\textit{MirrorDistill} is implemented in TensorFlow/Keras and trained on a single GPU. Images are processed with a resolution of $256\times256$, with random horizontal flipping and $90^{\circ}$ rotations as the only augmentations. We optimize with Adam (learning rate $2\times10^{-4}$, batch size $16$) for $1000$ epochs and clip gradients to a global
norm of $1.0$. The illumination weight uses $\beta=0.6$; the mirror losses use
$\lambda_{\text{enc}}=0.5$ and $\lambda_{\text{dec}}=0.8$; and the teacher decoder is EMA-updated every iteration with momentum $m=0.999$ (first initialized from the student weights). At inference, only the student autoencoder path is used while the clean-reference teacher path is used only during training and is removed at test time. Thus, the deployed model is the lightweight student encoder-decoder with no teacher-side inference cost. For evaluation, we report two standard full-reference metrics, \textit{Structural Similarity Index Measure} (SSIM), and \textit{Peak Signal-to-Noise Ratio} (PSNR).

\subsection{Results}
\subsubsection{Quantitative Results}
Table~\ref{tab:main_native} compares the proposed method with representative Retinex-based, CNN-based, GAN-based, and transformer-based enhancement methods on the LOL benchmarks. These methods include RetinexNet~\cite{Wei2018RetinexNet},
KinD~\cite{Zhang2019KinD}, EnlightenGAN~\cite{Jiang2021EnlightenGAN},
Restormer~\cite{Zamir2022Restormer}, MIRNet~\cite{Zamir2020MIRNet},
SNR-Aware~\cite{xu2022snr}, and Retinexformer~\cite{Cai2023Retinexformer}. As shown in the table, our method achieves 24.08 dB PSNR and 0.864 SSIM, giving the best performance among all compared methods on LOL-v2-real. This result is highly significant since LOL-v2-real contains \textit{real-world} low-light images rather than synthetically generated degradations, making it a useful benchmark for practical low-light sensing conditions. On LOL-v1, our method achieves the second-best SSIM of 0.842, while Retinexformer obtains the highest PSNR and SSIM on this set. On LOL-v2-syn, our method obtains the highest SSIM of 0.932 and the second-best PSNR of 25.07 dB. Overall, the results indicate that the proposed mirror-guided training is highly beneficial with real-world low-light data while maintaining competitive performance on the other benchmark splits.

\subsubsection{Computational Cost and Runtime}
The complexity results in Table~\ref{tab:main_native} show that our proposed approach also provides an efficient quality-complexity trade-off. It only requires 4.38 GMACs, which is 1-2 orders of magnitude lower than the compared methods. Fig.~\ref{fig:teaser} also shows this trend, where the proposed method is positioned in the upper-left region of the plot, corresponding to higher PSNR with lower computational cost. The measured inference time is 4.8 ms per image under our implementation setting, which is lower than the timed baselines reported in the table. Retinexformer has fewer parameters, but our method has a lower computational cost and shorter inference time. This behavior follows from the training design: the clean-reference teacher branch and mirror losses are used during optimization, while inference uses only the lightweight student network. Therefore, the proposed method improves performance on the real low-light benchmark without adding teacher-side computation during inference deployment.

In addition, Fig.~\ref{fig122} presents qualitative comparisons on LOL-v2-real for three
real-world-captured scenes. All methods substantially improve visibility over the
near-dark inputs. Consistent with the
quantitative results in Table~\ref{tab:main_native}, the proposed method closely matches the ground-truth reference in exposure and overall appearance while featuring the lowest computational cost among all compared methods (see Fig.~\ref{fig:teaser}), and while outperforming or being on-par with far heavier models in terms of image enhancement quality. 

\subsubsection{Ablation studies}
\label{ablation}
Table~\ref{tab:ablation} ablates our MirrorDistill' components and their impact on the LOL-v2-real image recovery quality at  $256{\times}256$ evaluation resolution. Starting from a reconstruction-only baseline 
at $21.81$\,dB, adding the encoder mirror and the decoder mirror individually raises
PSNR to $23.15$\,dB and $23.31$\,dB, and combining the two i.e., the full dual mirror, reaches $24.04$\,dB, a gain of $+2.23$\,dB and $+0.020$ SSIM over
reconstruction alone. The two mirrors are therefore complementary: the encoder mirror
aligns the low-light and clean representations early, while the decoder mirror
constrains the reconstruction trajectory, and neither alone matches their combination.
Removing the illumination-aware weighting by setting $\beta=0$ reduces PSNR to $22.81$ dB and SSIM to $0.9048$, suggesting that emphasizing darker regions is useful for the proposed distillation objective.


\begin{table}[]
\centering
\caption{Ablation of the proposed components and their impact on the LOL-v2-real image recovery quality at  $256{\times}256$
evaluation resolution. 
}
\label{tab:ablation}
\begin{tabular}{l cc}
\toprule
Configuration & PSNR & SSIM\\
\midrule
$\mathcal{L}_{\text{rec}}$ (\ref{eq:total}) only                       & 21.81 & 0.8888\\
\quad + encoder mirror                                & 23.15 & 0.9059\\
\quad + decoder mirror                                & 23.31 & 0.9014\\
\quad + both mirrors (MirrorDistill)                  & \textbf{24.04} & \textbf{0.9093}\\
\midrule
MirrorDistill without illumination weighting ($\beta{=}0$)         & 22.81 & 0.9048\\
\bottomrule
\end{tabular}
\end{table}

\section{Conclusion}
\label{sec:con}

This paper presented \textit{MirrorDistill}, an illumination-aware latent
distillation framework for efficient LLIE. Rather than
supervising only the final enhanced image, the proposed method guides the student
network with clean-domain latent targets at two depths: the raw encoder features and the standardized multi-scale decoder projections. The distillation is concentrated on under-exposed regions through an illumination-aware weighting, and an EMA clean teacher supplies stable training targets.
Because the teacher branch and the clean reference are used only during training,
inference relies solely on the lightweight U-Net student and incurs no additional
cost. Experimental results show that our proposed \textit{MirrorDistill} approach attains the highest PSNR and SSIM on the real-captured LOL-v2-real benchmark while requiring the lowest GFLPs among the compared methods, and that it remains competitive on LOL-v1 and LOL-v2-synthetic. Ablation studies confirm that the dual encoder--decoder mirror, and the illumination-aware weighting each contribute to the final performance. These findings indicate that training-time latent guidance can improve enhancement quality on real low-light data while preserving a compact, low-cost inference model, making the proposed method a promising candidate for industrial visual sensing applications such as nighttime surveillance, autonomous navigation, remote sensing, and inspection of poorly lit facilities.


\bibliographystyle{IEEEtran}
\bibliography{Refrences}

@article{Guo2016LIME,
  title={LIME: Low-Light Image Enhancement via Illumination Map Estimation},
  author={Xiaojie Guo and Yu Li and Haibin Ling},
  journal={IEEE Transactions on Image Processing},
  year={2016},
  volume={26},
  number={2},
  pages={982--993}
}

@article{Ying2017BIMEF,
  title={A Bio-inspired Multi-Exposure Fusion Framework for Low-light Image Enhancement},
  author={Zhenqiang Ying and Ge Li and Wen Gao},
  journal={arXiv preprint arXiv:1711.00591},
  year={2017}
}

@article{Li2021ZeroDCEpp,
  title={Learning to Enhance Low-Light Image via Zero-Reference Deep Curve Estimation},
  author={Chen Wei Li and Chongyi Li and Chen Change Loy},
  journal={IEEE Transactions on Pattern Analysis and Machine Intelligence},
  year={2021},
  volume={44},
  number={8},
  pages={4225--4238}
}

@article{Jiang2021EnlightenGAN,
  title={EnlightenGAN: Deep Light Enhancement without Paired Supervision},
  author={Yifan Jiang and Xinyue Gong and Ding Liu and Yu Cheng and Chen Fang and Xiaohui Shen and Jianchao Yang and Pan Zhou and Zhaowen Wang},
  journal={IEEE Transactions on Image Processing},
  year={2021},
  volume={30},
  pages={2340--2349}
}

@article{Wei2018RetinexNet,
  title={Deep Retinex Decomposition for Low-Light Enhancement},
  author={Chen Wei and Wenqi Ren and Wenhan Yang and Jiaying Liu},
  journal={arXiv preprint arXiv:1808.04560},
  year={2018}
}

@inproceedings{Zhang2019KinD,
  title={Kindling the Darkness: A Practical Low-light Image Enhancer},
  author={Yongjie Zhang and Jianqiang Zhang and Chao Xu and Xiaojie Guo},
  booktitle={ACM International Conference on Multimedia},
  year={2019},
  pages={1632--1640}
}

@inproceedings{Zamir2020MIRNet,
  title={Learning Enriched Features for Real Image Restoration and Enhancement},
  author={Syed Waqas Zamir and Aditya Arora and Salman Khan and Munawar Hayat and Fahad Shahbaz Khan and Ming-Hsuan Yang},
  booktitle={European Conference on Computer Vision},
  year={2020},
  pages={492--511}
}

@inproceedings{Wang2022UFormer,
  title={UFormer: A General U-shaped Transformer for Image Restoration},
  author={Zhendong Wang and Xiangyu Cun and Jianmin Bao and Wengang Zhou and Dong Chen and Fang Wen},
  booktitle={CVPR},
  year={2022},
  pages={17683--17693}
}

@inproceedings{Zamir2022Restormer,
  title={Restormer: Efficient Transformer for High-resolution Image Restoration},
  author={Syed Waqas Zamir et al.},
  booktitle={CVPR},
  year={2022},
  pages={5728--5739}
}

@inproceedings{Cai2023Retinexformer,
  title={Retinexformer: One-stage Retinex-based Transformer for Low-Light Image Enhancement},
  author={Yingkun Cai and Heng Bian and Jing Lin and Hao Wang and Radu Timofte and Yulun Zhang},
  booktitle={ICCV},
  year={2023},
  pages={12504--12513}
}

@article{Wu2025URetinexNetPP,
  title={Interpretable Optimization-inspired Unfolding Network for Low-light Image Enhancement},
  author={Weijian Wu and Junjie Weng and Peng Zhang and Xiaojie Wang and Wenqiang Yang and Jianbo Jiang},
  journal={IEEE Transactions on Pattern Analysis and Machine Intelligence},
  year={2025}
}

@inproceedings{jiang2024lightendiffusion,
  title={Lightendiffusion: Unsupervised low-light image enhancement with latent-retinex diffusion models},
  author={Jiang, Hai and Luo, Ao and Liu, Xiaohong and Han, Songchen and Liu, Shuaicheng},
  booktitle={European Conference on Computer Vision},
  pages={161--179},
  year={2024},
  organization={Springer}
}

@ARTICLE{yang2021sparse,
  author={Yang, Wenhan and Wang, Wenjing and Huang, Haofeng and Wang, Shiqi and Liu, Jiaying},
  journal={IEEE Transactions on Image Processing}, 
  title={Sparse Gradient Regularized Deep Retinex Network for Robust Low-Light Image Enhancement}, 
  year={2021},
  volume={30},
  number={},
  pages={2072-2086},
  doi={10.1109/TIP.2021.3050850}}

@inproceedings{woo2018cbam,
author = {Woo, Sanghyun and Park, Jongchan and Lee, Joon-Young and Kweon, In So},
title = {CBAM: Convolutional Block Attention Module},
year = {2018},
isbn = {978-3-030-01233-5},
publisher = {Springer-Verlag},
address = {Berlin, Heidelberg},
doi = {10.1007/978-3-030-01234-2_1},
booktitle = {Computer Vision – ECCV 2018: 15th European Conference, Munich, Germany, September 8–14, 2018, Proceedings, Part VII},
pages = {3–19},
numpages = {17},
location = {Munich, Germany}
}

@inproceedings{xu2022snr,
  title={Snr-aware low-light image enhancement},
  author={Xu, Xiaogang and Wang, Ruixing and Fu, Chi-Wing and Jia, Jiaya},
  booktitle={Proceedings of the IEEE/CVF conference on computer vision and pattern recognition},
  pages={17714--17724},
  year={2022}
}

@InProceedings{10.1007/978-3-031-73010-8_16,
author="Yu, Ye
and Chen, Fengxin
and Yu, Jun
and Kan, Zhen",
editor="Leonardis, Ale{\v{s}}
and Ricci, Elisa
and Roth, Stefan
and Russakovsky, Olga
and Sattler, Torsten
and Varol, G{\"u}l",
title="LMT-GP: Combined Latent Mean-Teacher and Gaussian Process for Semi-supervised Low-Light Image Enhancement",
booktitle="Computer Vision -- ECCV 2024",
year="2025",
publisher="Springer Nature Switzerland",
address="Cham",
pages="261--279",
isbn="978-3-031-73010-8"
}

@inproceedings{zhou2024unveiling,
  title={Unveiling advanced frequency disentanglement paradigm for low-light image enhancement},
  author={Zhou, Kun and Lin, Xinyu and Li, Wenbo and Xu, Xiaogang and Cai, Yuanhao and Liu, Zhonghang and Han, Xiaoguang and Lu, Jiangbo},
  booktitle={European Conference on Computer Vision},
  pages={204--221},
  year={2024},
  organization={Springer}
}

@inproceedings{hinton2015distilling,
  title={Distilling the Knowledge in a Neural Network},
  author={Hinton, Geoffrey and Vinyals, Oriol and Dean, Jeff},
  booktitle={NIPS Deep Learning and Representation Learning Workshop},
  year={2015}
}

@inproceedings{romero2015fitnets,
  title={FitNets: Hints for Thin Deep Nets},
  author={Romero, Adriana and Ballas, Nicolas and Kahou, Samira Ebrahimi and Chassang, Antoine and Gatta, Carlo and Bengio, Yoshua},
  booktitle={International Conference on Learning Representations},
  year={2015}
}

@inproceedings{zagoruyko2017attention,
  title={Paying More Attention to Attention: Improving the Performance of Convolutional Neural Networks via Attention Transfer},
  author={Zagoruyko, Sergey and Komodakis, Nikos},
  booktitle={International Conference on Learning Representations},
  year={2017}
}

@inproceedings{tarvainen2017mean,
  title={Mean Teachers are Better Role Models: Weight-Averaged Consistency Targets Improve Semi-Supervised Deep Learning Results},
  author={Tarvainen, Antti and Valpola, Harri},
  booktitle={Advances in Neural Information Processing Systems},
  year={2017}
}

@INPROCEEDINGS{10312449,
  author={Wang, Wei and Zheng, Chaobing},
  booktitle={IECON 2023- 49th Annual Conference of the IEEE Industrial Electronics Society}, 
  title={Unsupervised Low Light Enhancement Method with Inherent Diffuse Map}, 
  year={2023},
  volume={},
  number={},
  pages={1-6},
  doi={10.1109/IECON51785.2023.10312449}}

@INPROCEEDINGS{10905453,
  author={Wu, Yurui and Zhou, Shaodong and Zhang, Hui},
  booktitle={IECON 2024 - 50th Annual Conference of the IEEE Industrial Electronics Society}, 
  title={YOLOv8-LLE: An Improved Algorithm for Object Detection of Autonomous Driving Road Scenes in Low-Light Environments}, 
  year={2024},
  volume={},
  number={},
  pages={1-6},
  doi={10.1109/IECON55916.2024.10905453}}

@INPROCEEDINGS{9968642,
  author={Shireen Ansarnia, Masoomeh and Tisserand, Etienne and Tremeau, Alain and Schweitzer, Patrick},
  booktitle={IECON 2022 – 48th Annual Conference of the IEEE Industrial Electronics Society}, 
  title={Urban road users detection and velocity estimation from top-view fish-eye imagery under low light conditions}, 
  year={2022},
  volume={},
  number={},
  pages={1-6},
  doi={10.1109/IECON49645.2022.9968642}}

@article{Lore2017LLNet,
  author  = {Lore, Kin Gwn and Akintayo, Adedotun and Sarkar, Soumik},
  title   = {LLNet: A Deep Autoencoder Approach to Natural Low-Light Image Enhancement},
  journal = {Pattern Recognition},
  volume  = {61},
  pages   = {650--662},
  year    = {2017},
  doi     = {10.1016/j.patcog.2016.06.008}
}

@article{Li2022LLIESurvey,
  author  = {Li, Chongyi and Guo, Chunle and Han, Lin and Jiang, Jun and Cheng, Ming-Ming and Gu, Jinwei and Loy, Chen Change},
  title   = {Low-Light Image and Video Enhancement Using Deep Learning: A Survey},
  journal = {IEEE Transactions on Pattern Analysis and Machine Intelligence},
  volume  = {44},
  number  = {12},
  pages   = {9396--9416},
  year    = {2022},
  doi     = {10.1109/TPAMI.2021.3126387}
}

@misc{Liang2020DeepBilateralRetinex,
  author        = {Liang, Jinxiu and Xu, Yong and Quan, Yuhui and Wang, Jingwen and Ling, Haibin and Ji, Hui},
  title         = {Deep Bilateral Retinex for Low-Light Image Enhancement},
  year          = {2020},
  eprint        = {2007.02018},
  archivePrefix = {arXiv},
  primaryClass  = {eess.IV},
  doi           = {10.48550/arXiv.2007.02018}
}

@inproceedings{Zheng2022SGZ,
  author    = {Zheng, Shen and Gupta, Gaurav},
  title     = {Semantic-Guided Zero-Shot Learning for Low-Light Image/Video Enhancement},
  booktitle = {Proceedings of the IEEE/CVF Winter Conference on Applications of Computer Vision (WACV) Workshops},
  pages     = {581--590},
  month     = {January},
  year      = {2022}
}

@article{Lv2021DeepRetinexBilateral,
  author  = {Lv, Xiaoqian and Sun, Yujing and Zhang, Jun and Jiang, Feng and Zhang, Shengping},
  title   = {Low-Light Image Enhancement via Deep Retinex Decomposition and Bilateral Learning},
  journal = {Signal Processing: Image Communication},
  volume  = {99},
  pages   = {116466},
  year    = {2021},
  doi     = {10.1016/j.image.2021.116466}
}

@INPROCEEDINGS{10582024,
  author={Safa, Ali and Mommen, Wout and Wambacq, Piet and Keuninckx, Lars},
  booktitle={2024 IEEE 18th International Conference on Automatic Face and Gesture Recognition (FG)}, 
  title={Resource-Efficient Gesture Recognition Using Low-Resolution Thermal Camera via Spiking Neural Networks and Sparse Segmentation}, 
  year={2024},
  volume={},
  number={},
  pages={1-5},
  doi={10.1109/FG59268.2024.10582024}}

\end{document}